\def\year{2020}\relax
\documentclass[letterpaper]{article}
\usepackage{aaai20}
\usepackage{times}
\usepackage{helvet}
\usepackage{courier}
\usepackage{graphicx}
\usepackage{dirtytalk}
\usepackage{colortbl}

\usepackage{times}  
\usepackage{helvet} 
\usepackage{courier}  
\usepackage[hyphens]{url}  
\usepackage{graphicx} 
\usepackage{graphicx}  

\begin{document}
%

\setcounter{secnumdepth}{0}

\title{An Iterative Approach for Identifying Complaint Based Tweets in Social Media Platforms (Student Abstract)}
\author{Gyanesh Anand$^{1}$\thanks{Equal Contribution}, Akash Gautam$^{1}$\footnotemark[1], Puneet Mathur$^{2}$, Debanjan Mahata$^{3}$, \\ \bf \Large Rajiv Ratn Shah$^{1}$, Ramit Sawhney$^{4}$\\ 
$^{1}$ MIDAS, IIIT-Delhi \emph{\{gyanesh16039, akash15011, rajivratn\}@iiitd.ac.in}, \\
$^{2}$ University of Maryland, College Park \emph{puneetm@cs.umd.edu}, \\
$^{3}$ Bloomberg, New York, U.S.A. \emph{dmahata@bloomberg.net}, \\
$^{4}$ Netaji Subhas Institute of Technology \emph{ramits.co@nsit.net.in}\\
}

\maketitle
\begin{abstract}
\begin{quote}
Twitter is a social media platform where users express opinions over a variety of issues. Posts offering grievances or complaints can be utilized by private/ public organizations to improve their service and promptly gauge a low-cost assessment.
In this paper, we propose an iterative methodology which aims to identify complaint based posts pertaining to the transport domain. We perform comprehensive evaluations along with releasing a novel dataset for the research purposes\footnote{The dataset can be found at https://github.com/midas-research/transport-complaint-detection}. 

\end{quote}
\end{abstract}
\section{Introduction}
With the advent of social media platforms, increasing user base address their grievances over these platforms, in the form of complaints. According to \cite{olshtain1985complaints}, \textit{complaint is considered to be a basic speech act used to express negative mismatch between the expectation and reality.} Transportation and its related logistics industries are the backbones of every economy\footnote{https://www.entrepreneur.com/article/326552}. Many transport organizations rely on complaints gathered via these platforms to improve their services, hence understanding these are important for: (1) linguists to identify human expressions of criticism and (2) organizations to improve their query response time and address concerns effectively. 

Presence of inevitable noise, sparse content along with rephrased and structurally morphed instances of posts, make the task at hand difficult \cite{shah2017multimodal}. Previous works \cite{meinl2013electronic} in the domain of complaint extraction have focused on static datasets only. These are not robust to changes in the trends reflected, information flow and linguistic variations. \textbf{We propose an iterative, semi-supervised approach for identification of complaint based tweets, having the ability to be replicated for \textit{\say{stream of information flow}}}. The preference of a semi-supervised approach over supervised ones is due to the stated reasons: (a) the task of isolating the training set, make supervised tasks less attractive and impractical and (b) imbalance between the subjective and objective classes lead to poor performance.

\section{Proposed Methodology}
We aimed to mimic the presence of sparse/noisy content distribution, mandating the need to curate a novel dataset via specific lexicons. We scraped $500$ random posts from recognized transport forum\footnote{https://www.theverge.com/forums/transportation}. A pool of $50$ uni/bi-grams was created based on tf-idf representations, extracted from the posts, which was further pruned by annotators. Querying posts on \textit{Twitter} with extracted lexicons led to a collection of $19,300$ tweets. In order to have lexical diversity, we added $2500$ randomly sampled tweets to our dataset. In spite of the sparse nature of these posts, the lexical characteristics act as \textbf{information cues}. 

\begin{figure}[!t]
\centering
\includegraphics[scale=0.327]{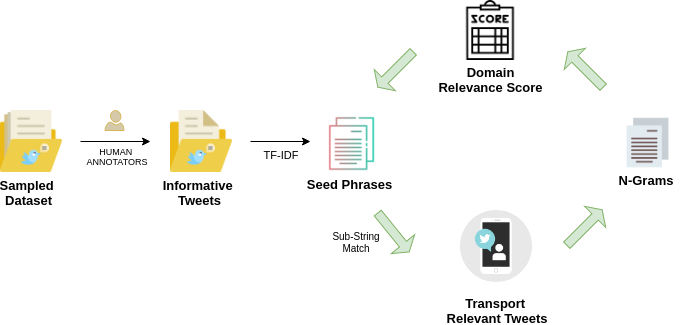}
\caption{Pictorial representation of the proposed pipeline.}
\label{fig:pipeline}
\end{figure}

Figure \ref{fig:pipeline} pictorially represents our methodology. Our approach required an initial set of \textbf{\textit{informative tweets}} for which we employed two human annotators annotating a random sub-sample of the original dataset. From the $1500$ samples, $326$ were marked as \textit{informative} and $1174$ as \textit{non informative} ($\kappa=0.81$), discriminated on this criteria: \textit{\textbf{Is the tweet addressing any complaint or raising grievances about modes of transport or services/ events associated with transportation such as traffic; public or private transport?}}. An example tweet marked as \textit{informative}: \textit{\say{No, metro fares will be reduced ???, but proper fare structure needs to presented right, it's bad !!!}}. 

We utilized tf-idf for the identification of initial \say{seed phrases} from the curated set of \textit{\textbf{informative tweets}}. $50$ terms having the highest tf-idf scores were passed through the complete dataset and based on sub-string match, the \textbf{\textit{transport relevant tweets}} were identified. The redundant tweets were filtered based on the cosine similarity score. \textbf{Implicit information indicators} were identified based on \textbf{\textit{domain relevance score}}, a metric used to gauge the coverage of n-gram ($1$,$2$,$3$) when evaluated against a randomly created pool of posts. 

We collected a pool of $5000$ randomly sampled tweets different from the data collection period. The rationale behind having such a metric was to discard commonly occurring n-grams normalized by random noise and include ones which are of lexical importance.  We used terms associated with high domain relevance score (threshold determined experimentally) as \say{seed phrases} for the next set of iterations. The growing dictionary augments the collection process. The process ran for $4$ iterations providing us $7200$ \textbf{\textit{transport relevant tweets }} as no new lexicons were identified. In order to identify linguistic signals associated with the \textit{complaint} posts, we randomly sampled a set of $2000$ tweets which was used as training set, manually annotated into distinct labels: \textit{complaint relevant} ($702$) and \textit{complaint non-relevant} ($1298$) ($\kappa=0.79$). We employed these features on our dataset.

\textbf{Linguistic markers}. To capture linguistic aspects of complaints, we utilized {\fontfamily{qcr}\selectfont Bag of Words}, count of {\fontfamily{qcr}\selectfont POS tags} and {\fontfamily{qcr}\selectfont Word2vec clusters}.

\textbf{Sentiment markers}. We used quantified score based on the ratio of tokens mentioned in the following lexicons: {\fontfamily{qcr}\selectfont MPQA}, {\fontfamily{qcr}\selectfont NRC}, {\fontfamily{qcr}\selectfont VADER} and {\fontfamily{qcr}\selectfont Stanford}.

\textbf{Information specific markers}. These account for a set of handcrafted features associated with \textit{complaint}, we used the stated markers 
(a) {\fontfamily{qcr}\selectfont Text-Meta Data}, this includes the count of URL's, hashtags, user mentions, special symbols and user mentions, used to enhance retweet impact;
(b) {\fontfamily{qcr}\selectfont Request Identification}, we employed the model presented in \cite{danescu2013no} to identify if a specific tweet assertion is a request; (c) {\fontfamily{qcr}\selectfont Intensifiers}, we make use of feature set derived from the number of words starting with capital letters and the repetition of special symbols (exclamation, questions marks) within the same post; (d) {\fontfamily{qcr}\selectfont Politeness Markers}, we utilize the politeness score of the tweet extracted from the model presented in \cite{danescu2013no}; (e) {\fontfamily{qcr}\selectfont Pronoun Variation}, these have the ability to reveal the personal involvement or intensify involvement. We utilize the frequency of pronoun types $\{\textit{first, second, third, demonstrative and indefinite}$\} using pre-defined dictionaries. 

From the pool of $7200$ transport relevant tweets, we sampled $3500$ tweets which were used as the testing set. The results are reported in Table\ref{tab:res} with $10$ fold cross-validation. \textbf{With increasing the number of iterations, the pool of \say{seed phrases} gets refined and augments the selection of \textit{\textbf{transport relevant tweets}}}. The proposed pipeline is tailored to identify complaint relevant tweets in a noisy scenario. 

\section{Results}
Table \ref{tab:res} reflects that the BOW model provided the best results, both in terms of accuracy and F1-score. The best result achieved by a sentiment model was the Stanford Sentiment ($0.63$ F1-score), with others within the same range and linguistic-based features collectively giving the best performance.

\definecolor{Gray}{gray}{0.9}

\begin{table}[]
\small
\centering
{\small \begin{tabular}{|c|cc|}
\hline
\textbf{Model}      & \textbf{Accuracy(\%)} & \textbf{F1-score} \\
\hline
\rowcolor{Gray}
\multicolumn{3}{|c|}{\textbf{Linguistic Markers}} \\
\hline
\textbf{Bag-of-Words} &\textbf{75.3}       &\textbf{0.71}       \\
\rowcolor{Gray}
POS Tags    &70.1       &0.66       \\
Word2Vec cluster &72.1   &0.67       \\
\hline
\rowcolor{Gray}
\multicolumn{3}{|c|}{\textbf{Sentiment Markers}} \\
\hline
Sentiment-MPQA  &68.2    &0.61       \\ 
\rowcolor{Gray}
Sentiment-NRC  &67.9     &0.59       \\ 
Sentiment-VADER &68.0    &0.62       \\
\rowcolor{Gray}
\textbf{Sentiment-Stanford} &\textbf{68.7} &\textbf{0.63}      \\
\hline
\multicolumn{3}{|c|}{\textbf{Information Specific Markers}} \\
\hline
\rowcolor{Gray}
Text Meta-Data  &69.3  &0.62     \\    
Request Identification  &70.1       &0.66           \\
\rowcolor{Gray}
\textbf{Intensifiers} & \textbf{72.5}  &\textbf{0.67}           \\
Politeness Markers   &70.4   &0.63   \\
\rowcolor{Gray}
Pronoun Variations    &69.6   &0.65       \\

\hline
\end{tabular}}
\caption{Performance of various linguistic, sentiment and information specific features on our dataset. \textbf{Classifier utilized Logistic Regression (Elastic Net regularization)}, as it gave the best performance as compared to its counterparts.}
\label{tab:res}
\end{table}

\section{Conclusion and Future Work}
In this paper, we presented a novel semi-supervised pipeline along with a novel dataset for identification of complaint based posts in the transport domain. \textbf{The proposed methodology can be expanded for other fields by altering the lexicons used for the creation of information cues}. There are limitations to this analysis; we do not use neural networks which mandate a large volume of data. In the future, we aim to identify demographic features for identification of complaint based posts on social media platforms.


\bibliography{ref}
\bibliographystyle{aaai}

\end{document}